\documentclass[conference]{IEEEtran}
\usepackage{CJKutf8}
\usepackage{comment}
\usepackage{amsmath}
\usepackage{tabularx}
\usepackage{booktabs}
\usepackage{multirow}
\usepackage{mathtools}
\usepackage{algorithm2e}
\usepackage{graphicx}
\usepackage{amssymb, mathtools}
\usepackage{capt-of}
\usepackage[pass]{geometry}
\usepackage{xcolor}
\usepackage{balance}
\usepackage{CJKutf8}
\input epsf
\usepackage{graphicx}

\begin{document}

%问题1 时序问题不像其他cv和NLP模型 模型输入有相对统一的数据结构 a unified framework for handling diverse tasks
%
% \title{Multi-weighted Graph Attention Networks for Passenger Count Prediction on Railway Network}
% \title{Multi-weighted Graphs Learning for Spatiotemporal Interploration for Passenger Count on the Traffic Network}Scalable Attention 
% \title{Multi-weighted Graphs Learning for Spatiotemporal Interpolation for Passenger Count on the Traffic Network}
\title{FRTP: Federating Route Search Records to Enhance Long-term Traffic Prediction}
% \title{Leveraging Online Route Search Records in Cyber Space for Improving Traffic Speed Prediction in Long Term}

% \title{Combining Route Search Records in Cyber Space with Spatio-temporal Features for Traffic Speed Prediction}

% \title{Spatial Interpolation based on Multi-weighted Graphs Learning on Scalable Neighboring}

\author{
    \IEEEauthorblockN{Hangli Ge\IEEEauthorrefmark{1}, Xiaojie Yang\IEEEauthorrefmark{2}, Itsuki Matsunaga\IEEEauthorrefmark{2}, Dizhi Huang\IEEEauthorrefmark{2}, Noboru Koshizuka\IEEEauthorrefmark{1}}
    \IEEEauthorblockA{
    \IEEEauthorrefmark{1}Interfaculty Initiative in Information Studies; }
    \IEEEauthorblockA{\IEEEauthorrefmark{2}Graduate School of Interdisciplinary Information Studies;
    \\The University of Tokyo} \{hangli.ge, xiaojie.yang, itsuki.matsunaga, dizhi.huang, noboru\}@koshizuka-lab.org
}
%Traffic forecasting is the core component of intelligent transportation systems (ITS).

\maketitle

\begin{abstract}
% Accurate traffic forecasting is crucial for intelligent transportation systems (ITS). In particular, forecasting traffic conditions several days in advance facilitates mid- and long-term traffic optimization. However, the inclusion of various external features alongside spatial domain and the uncertainty of temporal domain, increases the complexity of the models. Also, data preprocessing was always handled separately from the learning model, leading to inefficiencies due to frequent trial repetition of preprocessing and training.
% In this research, we propose a federation architecture for learning which is able to learn from raw data with varying features and time granularity. The model employs a unified design capable of learning from  different feature types, time granularities and time periods. In the experiments, our proposed model utilizes route search records, starting with the design of processing raw data. Unlike previous models, this research integrates the data preprocessing phase into the learning process, making the model compatible with various time frequencies and input/output lengths. The applicability of the proposed model is demonstrated through evaluations of different learning patterns of parameter settings.

Accurate traffic prediction, especially predicting traffic conditions several days in advance is essential for intelligent transportation systems (ITS). Such predictions enable mid- and long-term traffic optimization, which is crucial for efficient transportation planning. However, the inclusion of diverse external features, alongside the complexities of spatial relationships and temporal uncertainties, significantly increases the complexity of forecasting models. Additionally, traditional approaches have handled data preprocessing separately from the learning model, leading to inefficiencies caused by repeated trials of preprocessing and training. In this study, we propose a federated architecture capable of learning directly from raw data with varying features and time granularities or lengths. The model adopts a unified design that accommodates different feature types, time scales, and temporal periods. Our experiments focus on federating route search records and begin by processing raw data within the model framework. Unlike traditional models, this approach integrates the data federation phase into the learning process, enabling compatibility with various time frequencies and input/output configurations. The accuracy of the proposed model is demonstrated through evaluations using diverse learning patterns and parameter settings. The results show that online search log data is useful for forecasting long-term traffic, highlighting the model's adaptability and efficiency.

% e utilize online search records for capturing the users' actions or intentions in cyberspace, which could be valuable for improving long-term traffic prediction.

% In addition, reduce practicality. Our proposal first leverages the route search records in cyber space for improving traffic speed. 

% In recent years, various deep learning models have been proposed for the prediction problems in traffic domain. However, most existing studies focused on short-term predictions in which timely result was output. In addition,and at a macro level

% traditional methods cannot satisfy the requirements of mid-and-long term prediction tasks and often neglect spatial and temporal dependencies.It makes the tasks not suitable for travel planning a few days in advance.

\end{abstract}

% To achieve great performance of prediction as well as to discover the correlation between learning model performance and feature processing. that compared with LSTM which is the best among three baselines,

\begin{keywords}
Traffic Prediction; Intelligent Transportation System; Online Search Record; Federation Architecture;
\end{keywords}

 %and compute on the features adequately 
 
\section{Introduction}
Traffic forecasting is one of the most essential components of intelligent transportation systems (ITS). It facilitates many applications including traffic control, route guidance, congestion avoidance, dynamic pricing, etc. For instance, forecasting the traffic in long-term enables more flexible and intelligent travel plans, it further  decreases traffic congestion and accidents, finally improves both time and economic efficiency by promoting dynamic pricing, and so on.

With the increasing availability of big traffic data, data-driven traffic prediction methods in ITS have shown considerable promise in recent years. However, the majority of these tasks focused on short-term prediction in which the timely result of several minutes later (e.g., 5 minutes or 60 minutes) is predicted. Long-term prediction, for instance, predicting the traffic conditions of one week ahead remains a challenging task. Additionally, when applying models across varying feature types, time granularities and time periods, there is a lack of unified model design that can feasibly federate all the features.

% There is a lack of interpretability in identifying the importance of variables in predicting traffic condition.

Various deep learning methods, such as LSTM (long short-term memory) \cite{tian2015predicting,ma2015long}, CNN (convolutional neural networks) \cite{yang2019mf,ma2017learning} and GCN (graph convolutional networks) \cite{ye2020build} have been proposed to learn spatio-temporal dependencies in the traffic domain \cite{yin2021deep}. However, previous methods based solely on historical traffic pattern recognition are limited in accuracy. Since traffic status is influenced by various external factors, incorporating these features can improve prediction accuracy. On the other hand, this also increases computational complexity. Several previous studies incorporated external features such as calendar information \cite{zhang2018predicting}, weather data \cite{ryu2019intelligent}, and event data such as traffic accidents and road construction \cite{zhang2017deep}. While most research begins with data that has already been collected and preprocessed, few research proposals address the integration of heterogeneous raw data from a foundational perspective.

%The proposed methods are primarily reliant on historical traffic data using neural networks such as LSTM and CNN \cite{zang2018long}. Predictions based merely on historical traffic records are limited in accuracy, 

% Additionally, research using map queries \cite{liao2018deep}, has only targeted locations around hot spots for prediction, making it difficult to apply the prediction results with geographical versatility. Nevertheless, collecting and integrating corresponding data for diverse contexts is challenging. 

\begin{figure}[t]
\centering
\includegraphics[height=1.75in, width=1\linewidth]{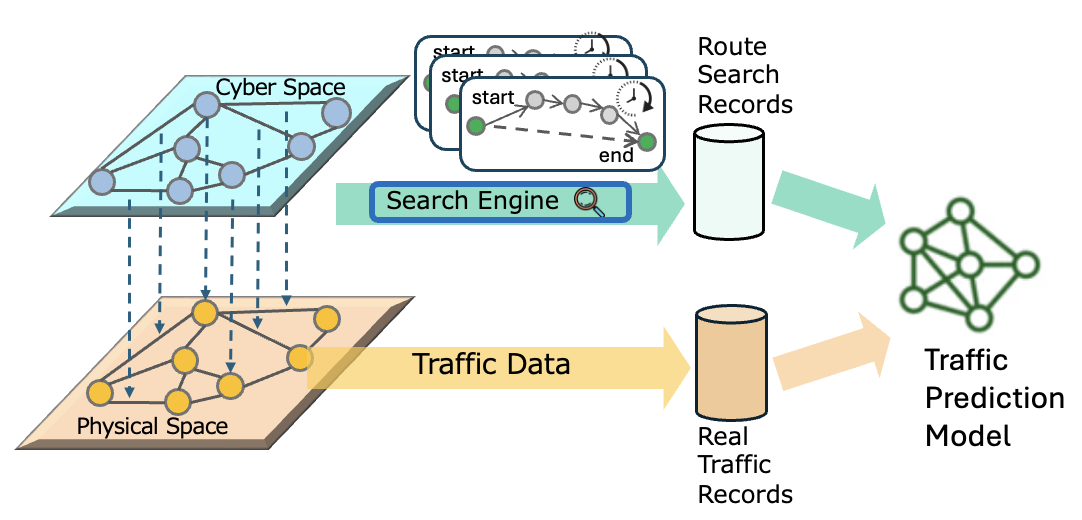}
\caption{Data federation architecture of our proposal}
\label{fig:sv}
\end{figure}

In this research, we propose a federated architecture of fusing route search records and traffic data for improving long-term prediction accuracy. As a practical solution, we introduce not only the training model, but also methods for data preprocessing, feature engineering, and more. Unlike other similar works, we utilized the most recent data from East Nippon Expressway Co., Ltd. (NEXCO East) \cite{nexco} for evaluation. NEXCO provides toll and route search service nationwide in Japan. As illustrated in Fig.~\ref{fig:sv}, this research presents a data process architecture that integrates route search records in cyberspace with traffic data in physical space. We proposed a method for processing raw route search records into a structured format that reflects potential real-world traffic volumes.

% Based on the graph clustering idea, we proposed Cluster-GCN, an algorithm to design the batches based on efficient graph clustering algorithms (e.g., METIS [8]). We take this idea further by proposing a stochastic multi-clustering framework to im- prove the convergence of Cluster-GCN. Our strategy leads to huge memory and computational benefits. In terms of memory, we only need to store the node embeddings within the current batch, which is O(bFL) with the batch size b. This is significantly better than VR-GCN and full gradient decent, and slightly better than other SGD-based approaches. In terms of computational complexity, our algorithm achieves the same time cost per epoch with gradient descent and is much faster than neighborhood searching approaches. In terms of the convergence speed, our algorithm is competitive with other SGD-based approaches. Finally, our algorithm is simple to implement since we only compute matrix multiplication and no neighborhood sampling is needed. Therefore for Cluster-GCN, we have [memory: good; time per epoch: good; convergence: good]. Heterogeneity

% \textbf{Previous traffic prediction}

Therefore, the contributions of this research are as follows:
\begin{itemize}
\item We propose a federation architecture which is \textbf{adaptive to learning features with various spatial and temporal dimensions}. In this proposal, setup of the model is simplified, requiring only a few parameters to be assigned. In addition, the model can compute different types of features, demonstrating its applicability. It can also handle various types of input data with different periodic granularities, reducing the labor effort required for data preprocessing and feature engineering.
\item We verified the effectiveness of various feature types focused on \textbf{long-term} prediction. In our research, in addition to the internal features (for instance, the road structure, real-time traffic data, etc.), other external data, i.e., search log data has been fully utilized. In particular, the search log data provided by our collaborated company which is rare and precious, has been utilized and evaluated for the practical experiment. The evaluation results demonstrate that \textbf{the volume of time-specified route searches has a strong correlation with future traffic volumes, making it valuable for long-term traffic prediction.} 
\end{itemize}

\section{Related work}

\subsection{Previous Learning Models}
Previous methods for traffic prediction can be categorized into three types: classical statistical methods, machine learning methods, and deep learning methods \cite{yin2021deep}. In the early age, statistical methods including Kalman filtering \cite{okutani1984dynamic}, ARIMA \cite{williams2003modeling} and VAR \cite{chandra2009predictions}, and so on were prevalent because the stationarity-based mathematical theories are solid.
However, these methods lacking flexibility are weak in traffic data that are highly non-linear and dynamic, resulting in poor performance in practice. Subsequently, traditional machine learning approaches such as Random Forest (RF) \cite{leshem2007traffic}, K-Nearest Neighbors (KNN)\cite{may2008vector}, Support Vector Machine (SVM)\cite{fu2016vehicle} have been proposed as improvements. The machine learning methods are able to model non-linearity and extract more complex feature-correlations in traffic data, further learn flexibly for better performance.

In recent years, owing to the increasing volume of big data in the traffic domain as well as the improved computational power, deep learning methods have been widely employed for traffic prediction with high prediction accuracy. Among deep learning methods, Recurrent neural network (RNN) \cite{ma2015long}, Long short term memory network (LSTM)\cite{zhao2017lstm} are the most widely used because of their advantage in capturing temporal dependencies. On the other hand, Convolutional Neural Network (CNN) \cite{ma2017learning} becomes more popular in modeling and learning spatial dependencies. Also, a number of hybrid models that combine CNN and LSTM have been proposed to capture complex spatio-temporal dependencies in traffic data \cite{wu2016short,li2017diffusion,cheng2018deeptransport}. 
In particular, Graph Convolutional Networks (GCNs) excel at capturing the non-Euclidean characteristics of traffic networks, making them highly effective for traffic prediction tasks
\cite{ge2024k,hangli2022multi}.

\subsection{Data Sources}
Sensor measurement datasets are prevalently utilized in existing works. Traffic information (e.g., traffic speed, amount, and occupancy) is generally collected during a short time interval by the detectors installed on the road network. There are several versions of open data in metropolises globally \cite{yin2021deep}, such as PEMS (Performance Measurement System) data with its sub-versions, including PEM3/4/7/8 and PEMSD-BAY in California \cite{pems},  Q-traffic in Beijing \cite{q-traffic}, LOOP dataset in Greater Seattle Area \cite{seattle-loop}, and so on. These datasets are extremely valuable and have been widely used and further promote a series of excellent research. However, the datasets have limitations such as the collection periods, sensed area or scales, lacking correspondently external features which narrows the practicality and contributions of the research. To further improve the accuracy, proposals of incorporating external data, including calendar information \cite{zhang2018predicting}, weather conditions: weather state (sunny/rainy/windy/cloudy, etc.), temperature, humidity, visibility \cite{yang2019mf}, road construction attributes \cite{ryu2019intelligent}, search records \cite{kosugi2022traffic,matsunaga2023improving} have been proposed.

\subsection{Spatio-temporal Dependencies for Long-term Prediction}

% This problem is challenging mainly due to the complex spatial and temporal dependencies [19]. 
Traffic time series generally demonstrate strong temporal dynamics. Several prediction targets, such as average speed, amount of vehicles, congestion status, travel time and so on, have been proposed. In normal situations, the traffic patterns may be cyclically recursive. On the other hand, abnormal traffic situations, for instance, congestion or accidents can result in the formation of non-stationary time series, rendering forecasts challenging. In particular, aiming for long-term predictions, the problem is more challenging because the correlation of the future statuses is difficult to capture. It is the reason that most of the existing studies have been conducted for traffic predicting of the next 5 min to 1h. Little research has been conducted to predict long-term, e.g., one day or one week later, because it is more challenging.

To summarize, long-term traffic prediction in practice presents several challenges, including: (1) developing a federation architecture that accommodates diverse feature data, and (2) effectively evaluating the impact of various input features. Notably, few studies have proposed methods for integrating input features with differing time granularities and lengths, which remains a significant gap in this research area. This motivates us to propose FLTP: a federated approach utilizing route search records to improve long-term traffic prediction. 

\section{Problem Definition and Notations}

\subsection{Formulation and Optimization Target} 
The traffic of stations was considered to be a time-series. Suppose that we have $T$ time slices, traffic can be denoted as $\mathcal{P} \in \mathbb{R}^{N\times T}$, and each road section has the history data of [$P_{0},P_{1},...,P_{t-1}$], and $N$ is the number of road segments and $T$ is the length of all the historical data. The problem is to predict the traffic status at the next time ($\hat{P_{t}}$) based on the historical data with the length of $T-1$. 

Hence, the prediction problem can be defined according to Equation~\ref{eqn:fun}:
\begin{equation}
\hat{P_{t}}=\mathit{f}(X_{t},X_{s},X_{r},P_{t-n},...,P_{t-2},P_{t-1})
\label{eqn:fun}
\end{equation}
where $\mathit{f}$(·) is the mapping function to be learned using the proposed deep-learning model. In addition, the optimization target is aiming to: minimize  $\parallel P_{t} - \hat{P_{t}} \parallel$ where the input of $X_{t} \in \mathbb{R}^{N\times T_{1}}$ is the  feature matrix of historical traffic data with the time length of $T_{1}$, $X_{s} \in \mathbb{R}^{N\times T_{2}}$ denotes the feature matrix of historical search records with the time length of $T_{2}$ and $X_{r} \in \mathbb{R}^{N\times 1}$ denotes the feature of road segments. $\hat{P_{t}}$ is the predicted traffic volume, $P_{t}$ is the ground truth. Notice that $T_{1}$ would not be equal to $T_{2}$. Therefore, in this research, learning on data of different lengths increases the difficulty.

% However, differentiate with previous research, the problem is historical $F_{t}$ 

\section{Datasets and Pre-processing}
With the collaboration of NEXCO East, two main types of datasets are provided: (1) traffic counter data (denoted as $X_{t}$ in equation~\ref{eqn:fun}) in which the road network data (denoted as $X_{r}$) was included; (2) online route search records (denoted as $X_{s}$) collected by a route and toll search engine called `Drive Plaza' \cite{dorapura}.

% 2) time specified online search log data; 3) time unspecified online search log data.

\subsection{Traffic Counter Data}
The traffic counter data is real-time sensed on the road. It includes the number of vehicles passing through, with the values of average speed and occupancy, etc., are recorded on each road section. The data is uniquely identified by the section ID and timestamp. The timestamp indicates the time when the data was measured. The traffic counter data was originally collected for each five minutes. We utilized the data of three expressways with the road codes of E4, E14 and E17 for the experiment and evaluation. The road network data is static and is used to describe the connection of road sections.

% As shown in Fig.~\ref{fig:qv}, a so-called Q-V (Quantity-velocity) diagram with the plotted number of vehicles on the horizontal axis and velocity on the vertical axis. It can be seen that the Q-V values could be separated in mainly two clusters: free flow and congestion flow. 

% How traffic congestion is evaluated can significantly affect traffic analysis and prediction and transport planning and applying potential congestion reduction strategies. In addition, unlike discrete sampling of a single feature such as speed or count, congestion is an event that needs to be measured by multiple features, i.g., speed, count, occupancy and so on. Also, it is generally expressed in spatiotemporal dimensions with a start and an end showing the distance and time length. Moreover, depend on the roadway condition such as the limited speed, lane count, etc, the definition of congestion varies. 

% In this study, we use the original traffic counter data for training to predict the speed prediction in 1-hour frequency. 

% Therefore, this is an appropriate area to validate that incorporating online search log data improves accuracy. The past T steps of historical traffic data are formatted into a tensor X∈RT×|E|×d(Xbefore input to the prediction model.

% \begin{figure}[h]
% \centering
% \includegraphics[height=1.8in, width=0.7\linewidth]{img/road_kepler.png}
% \caption{The targeted roads in this study}
% \label{fig:kepler}
% \end{figure}

\subsection{Online Route Search Records}
Online search records are collected through the online search service called `Drive Plaza' \cite{dorapura}, which is an expressway toll and route search service provided by NEXCO East. Each raw search record consists of five fields: $\textless$departure IC, arrival IC, departure time, arrival time, search time$\textgreater$, in which the values of departure IC, arrival IC and search time are required to be specified and other variables setting are optional. 

% We used only queries that searched for routes passing through the two roads.

Based on whether the user specified a depature or arrival datetime, we divided the search records into two types: \textbf{time-specified search logs} and \textbf{non-time-specified search logs}. The process of search records is as follows. 

\begin{enumerate}
\item Firstly, for all the route search records, we extracted the IC lists showing the shortest route between departure IC and arrival IC. For simplicity, the Dijkstra’s method was used as the shortest route extraction method.
\item If either departure time or arrival time was specified, we calculate the expected passage time for each IC segment within the route based on the specified datetime. We assume an average travel speed of 80 km/h on expressways, which is consistent with the speed limit on many expressways in Japan. 
\item If neither departure time nor arrival time was specified, we calculated the search records based on search time.
\item Since the search records are irregular on the time line that is unsuitable as input for the model. We originally sampled the search records by 5 minutes. If a search record exists, 1 will be added to the values of all the IC pairs from the departure to arrival.
\end{enumerate}
\begin{figure}[h]
\centering
\includegraphics[height=1.3in, width=1\linewidth]{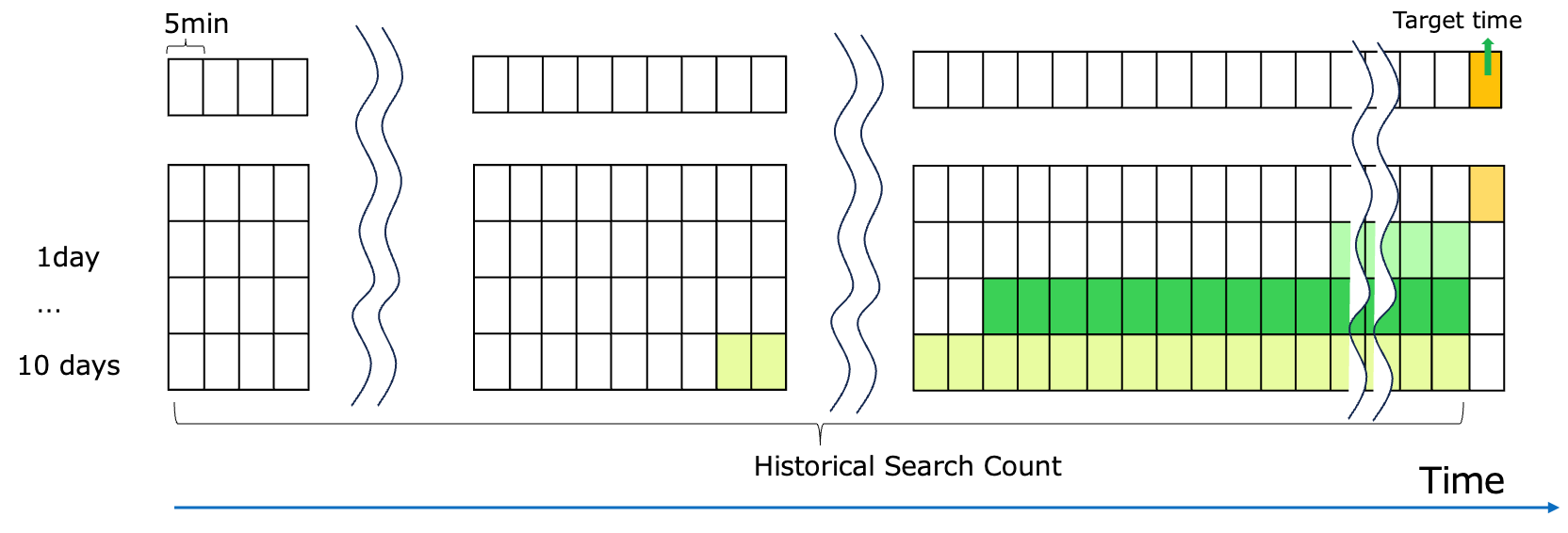}
\caption{Various resample patterns on the non-time-specified search records}
\label{fig:resample}
\end{figure}

Because drivers may search for routes by specifying departure or arrival time, time-specified search records suggest future traffic demands more directly. However, non-time-specified search records may have also some influence on real-world traffic. It is challenging to determine at what time point in the future the search records will impact traffic conditions. Therefore, as shown in Fig.~\ref{fig:resample}, regarding the non-time-specified search records, various patterns were calculated. In this study, we resampled a total of four patterns, from 1 day, 3 days, 7 days, and 10 days before the target time to be predicted.

\subsection{Analysis of Route Search Records}
\begin{figure}[h]
\centering
\includegraphics[height=1.5in, width=1\linewidth]{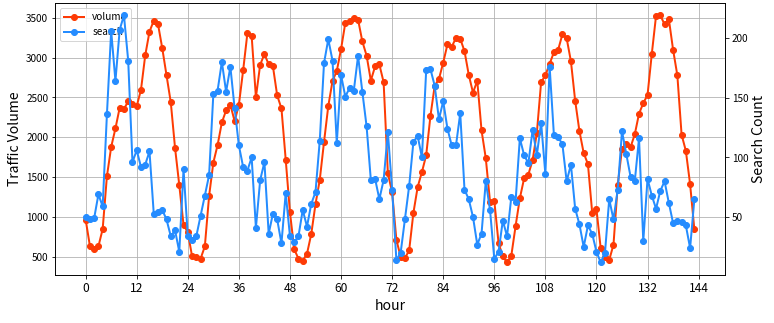}
\caption{Traffic volume and online search log data at 1-hour intervals. Online search log data captures the trend of actual traffic volume growth}
\label{fig:corre}
\end{figure}

\begin{figure}[h]
\centering
\includegraphics[height=1.3in, width=1\linewidth]{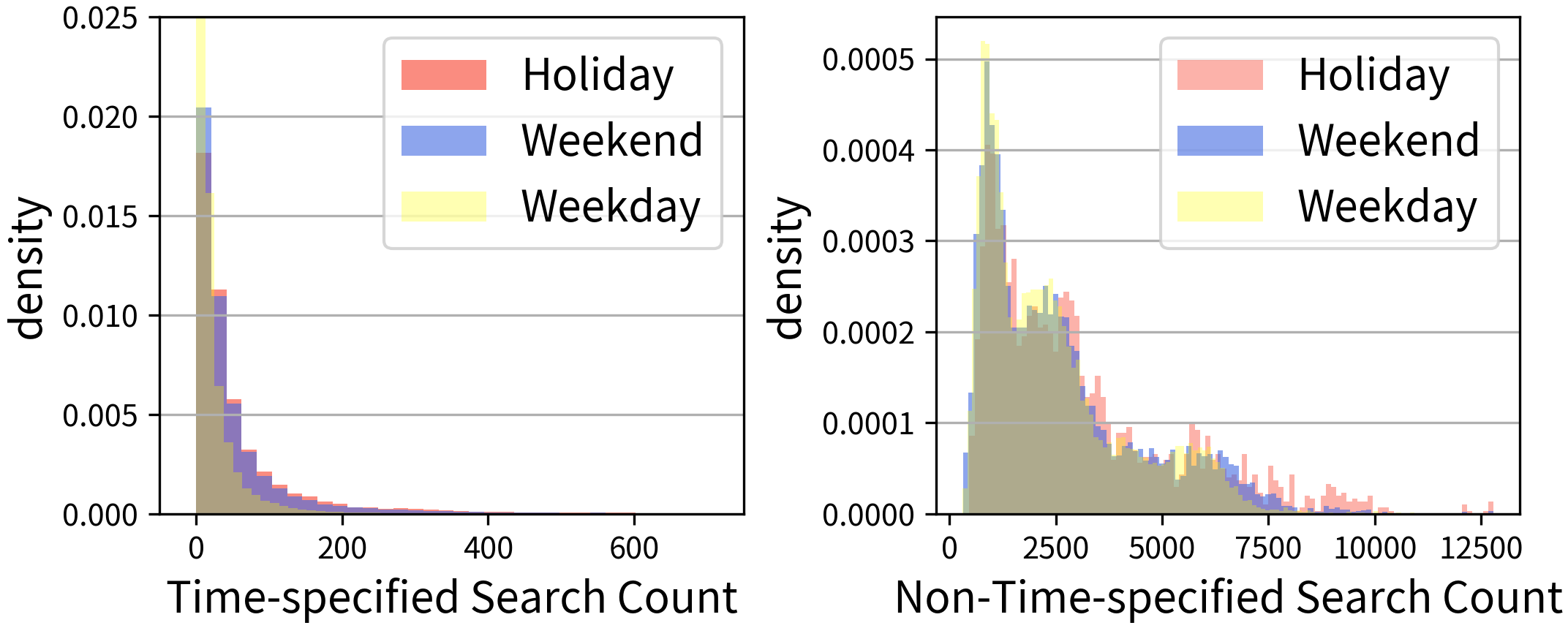}
\caption{The comparison of the distributions of two types of search counts}
\label{fig:ana1}
\end{figure}

\begin{figure}[h]
\centering
\includegraphics[height=3.3in, width=1\linewidth]{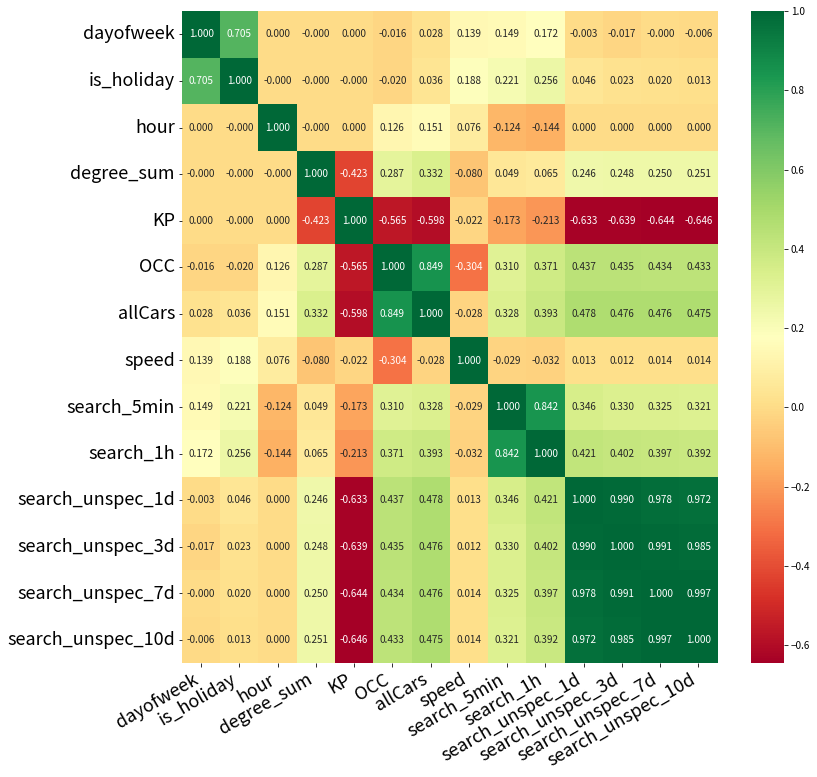}
\caption{The correlation of various features}
\label{fig:heatmap}
\end{figure}
As shown in the example in Fig.~\ref{fig:corre}, in which the historical traffic data and the structured time-specified search log data are resampled at one-hour intervals, the fluctuations in traffic volume and search volume closely align, capturing very similar behaviors. It demonstrates that the time-specified search log data captures the trend of actual traffic volume growth, especially during the morning hours.

\begin{figure*}[h]
\centering
\includegraphics[height=2.5in, width=1\linewidth]{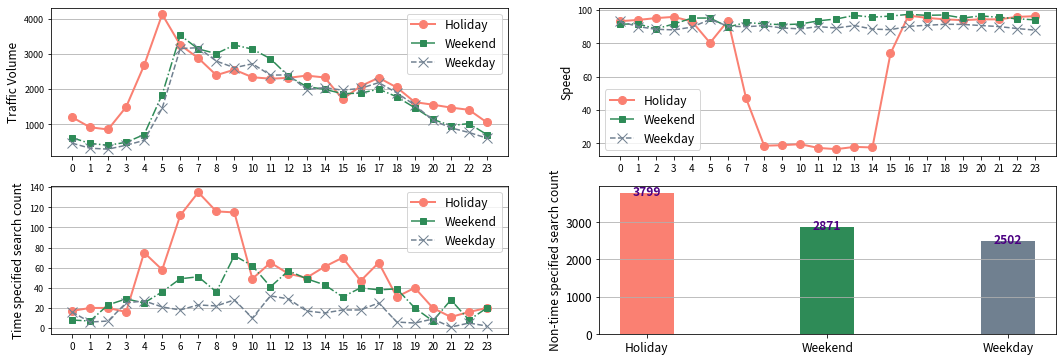}
\caption{Traffic Volume (upper left), time-specified search count (lower left), speed (upper right) and non-time-specified search count (lower right) for the specific segment between Miyoshi and Kawagoe of E17. The red line corresponds to a holiday (May 3, 2022), the green line to a weekend (May 14, 2022), and the gray line to a weekday (May 9, 2022).}
\label{fig:ana4}
\end{figure*}
% Thus, we use both search logs for traffic prediction.predictions was counted. 
% The historical traffic data and the structured online search log data are resampled at one-hour intervals.Long-term prediction results at one-hour intervals would be sufficient to support traffic management decision-making.

% Finally, we resampled the search records by different time intervals. 
% \begin{enumerate}
% \item With Time Specification: we convert the logs into potential future traffic volumes between interchanges (ICs) on the shortest route from departure to arrival.
% \item Without Time Specification: It is challenging to determine at what future point in time the search records will impact traffic conditions.
% \end{enumerate}

% \subsection{Data Aggregation}
% The region and time period included in the data  are consistent with the historical traffic data. This dataset contains approximately 16.5 million online search queries. Table I shows some examples of queries. 

% This data is structured spatiotemporally, and utilized as a feature indicating potential future traffic volume in the form of a tensor.
Fig.~\ref{fig:ana1} presents the distribution of
the time-specified (left) and the non-time-specified (right) search count for weekdays, weekends, and holidays. The data shows that more searches are conducted on holidays and weekends compared to weekdays. This is likely because weekday travel is often work-related, whereas on holidays, people are more likely to travel for leisure to unfamiliar destinations, prompting them to search for routes in advance.

Fig.~\ref{fig:heatmap} presents a correlation matrix showing relationships between different variables such as day-of-week, is\_holiday, hour, search count related feature, and others. We extracted several features which have correlation:

\begin{itemize}
\item \textbf{dayofweek}: Weakly correlated with search variables like search\_5min (0.149), search\_1h (0.172).
\item \textbf{is\_holiday}: Moderate positive correlation with search-related variables (0.22 to 0.25), indicating higher search activity around holidays.
\item \textbf{degree\_sum}: Positively correlated with search variables, especially longer periods like search\_unspec\_10d (0.251).
\item \textbf{KP} (Kilo Post segment, approximately 2 kilometers: Strong negative correlation with search variables (-0.63 to -0.65). As the KP value increases, indicating a location farther from the city center, the search values gradually decrease, leading to a negative correlation coefficient.
% Because the larger the KP value, the farther the location is from the city center, which leads to a gradual decrease in search values, finally leads to negative correlation coefficient.
\item \textbf{OCC} (Occupancy): Strong positive correlations with search variables (0.31 to 0.44), suggest that higher search values are associated with increased car presence and occupancy levels.
\item \textbf{allCars}: Similar to the value of OCC, moderate correlation with search variables (0.33to 0.48).
\item \textbf{speed}: Negatively correlated with search variables (-0.03) and moderately with OCC (-0.30), meaning speed decreases with higher occupancy.
\item \textbf{search count} variables (i.g., search\_5min, search\_1h, search\_unspec\_1d, etc.): Strong positive intercorrelations, with search activity across different time windows being highly correlated (up to 0.997 between search\_unspec\_7d and search\_unspec\_10d). Moderate positive correlation with OCC and allCars, suggesting that as occupancy and car counts increase, search activity increases as well.
\end{itemize}

Similarly, Fig.~\ref{fig:ana4} illustrates traffic volume, speed, time-specified search count, and non-time-specified search count for a specific road segment across a weekday, weekend, and holiday. On weekday, there is minimal correlation between the time-specified search count and traffic volume or speed. However, on weekend and holiday, the search count data correlates with the future traffic trends, such as the rapid increase in traffic volume and the sudden drop in speed in the morning. The non-time-specified search count follows an ascending pattern from weekday to weekend to holiday, corresponding closely with overall traffic volume trends, suggesting it provides insight into future traffic conditions. 

In summary, key insights include high correlations between search activity and variables like OCC and allCars, while KP negatively impacts many variables.  This analysis suggests that structured online search records hold potential in prediction of future traffic demand. During periods of high online search activity, such as weekends and holidays, these records offer a more detailed projection of upcoming traffic conditions. Consequently, \textbf{online search log data appear especially useful in forecasting unpredictable congestion, particularly during periods of unexpectedly high demand}.

\section{Proposed Model}
In this research, we propose a federated architecture of novel training model that effectively exploits and learns on three-dimensional dependencies: spatial, temporal, and cyberspace dimensions. This approach differs from previous proposals, which relied on downsampling techniques to compress data in the temporal dimension. Our proposed federation architecture is designed to (1) process raw data with varying time granularities; and (2) enhance the long-term prediction accuracy of future traffic.

\subsection{Convolutional LSTM based Learning Model}

% Together with utilizing the online search logs, the prediction has been improved.
\begin{figure*}[h]
\centering
\includegraphics[height=2.6in, width=0.7\linewidth]{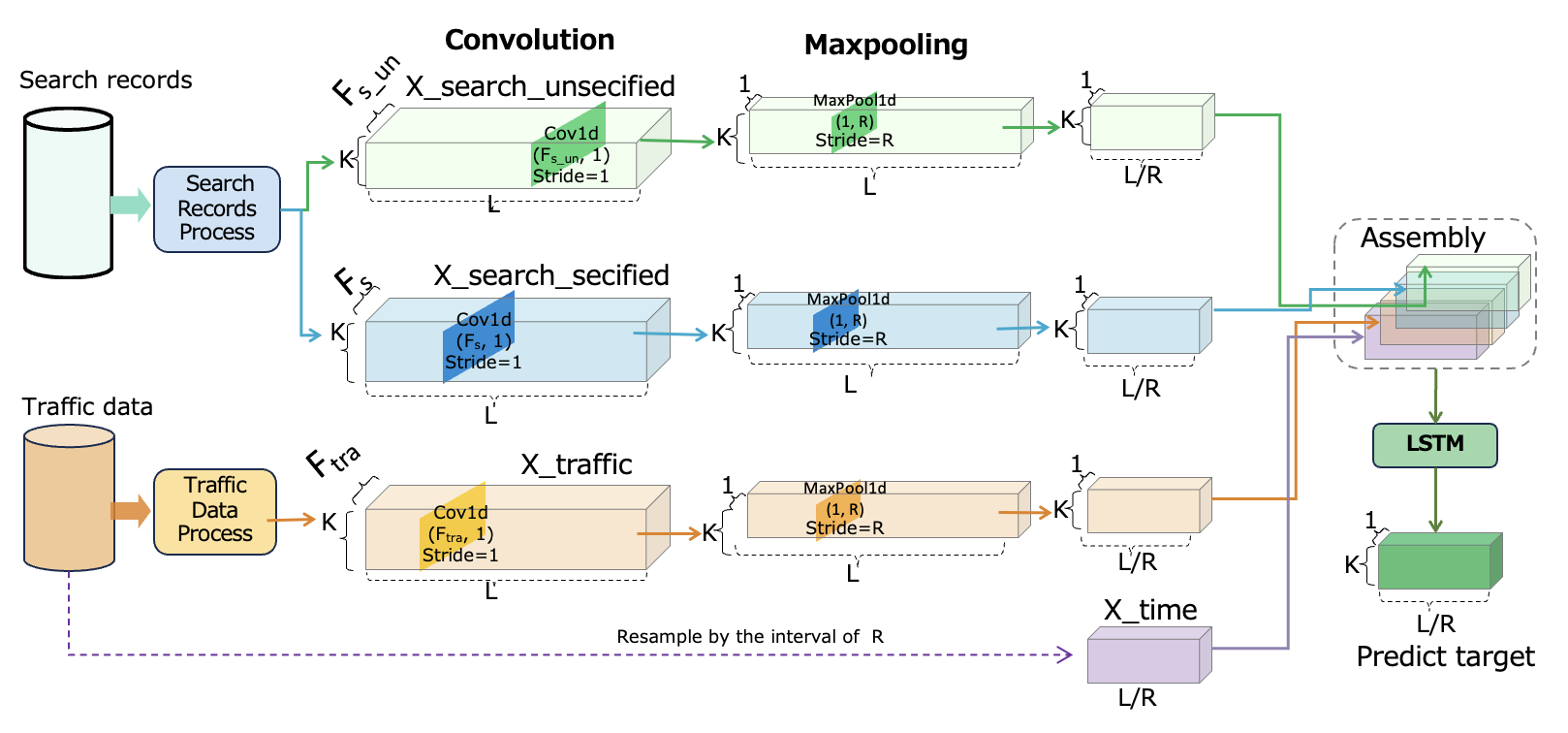}
\caption{The proposed model}
\label{fig:model}
\end{figure*}

%  K&  
% L& 
% R &   $F_{s\_un}$, $F_{s}$,$F_{tra}$
% F\_&    
% F\_&  
% F\_&   

Our proposal is a combination of CNN and LSTM as the convolutional LSTM architecture. As illustrated in Fig.~\ref{fig:model}, our proposed model consists of convolution, maxpooling and LSTM (Long short-term memory) layers.
The convolution operates on the feature dimensions, while the max-pooling layer and the LSTM work on the time dimension.
Table~\ref{tab:modelPara} lists the related parameters of the proposed model.
\begin{table}[]
\centering
\caption{Description of parameters}
\begin{tabular}{l|l}
\toprule
Parameter   & Description \\ \hline
 K&  The number of road segments \\
 L&  The time length of the training data\\ 
 R&   The ratio of input data length and output data length \\
 $F_{s\_un}$& The number of features of non-time specified search records \\ 
$F_{s}$ &  The number of features of time specified search records\\
$F_{tra}$ & The number of features of traffic data \\
 \bottomrule
\end{tabular}
\label{tab:modelPara}
\end{table}

\textbf{Convolution network:} We first applied the convolution network, which is useful to handle datasets with autocorrelated observations. Rather than using a fixed sample rate for resampling data, the convolution layer provides to extract the most effective data, and more parameters can be learned flexibly.

As shown in Fig.~\ref{fig:model}, the convolutional layer operates on the feature dimensions, which consist of three types. The number of features in each feature dimension is abstracted into different parameters, e.g., $F_{s\_un}$, $F_{s}$, $F_{tra}$ with the detaled description listed in Table \ref{tab:modelPara}. Regardless of the parameter value for the number of features, the convolutional layer will ultimately compress the input features into output features with a fixed number. In the experiment, we ultimately fixed the value at 1, as it achieved relatively strong performance while maintaining simplicity.

\textbf{Max-pooling layer:} Max-pooling layers were applied to the feature data by extracting the maximum value within each interval of R. Additionally, the max-pooling layer operates with a stride of R, meaning it extracts the maximum value in each time interval of R. Also, regardless of how the input or output length scale changes, R is an encapsulated variable and can be used in all cases. For example, if the input data was 5-minute time granularity and the output was 1 hour, R would be 12 (=60/5). If the input was a 15-minute time granularity and the output was 1 hour, R would be 4 (=60/15). Our proposed model is able to accommodate patterns across all varying time granularities.

\textbf{LSTM (Long Short-Term Memory):} is a recurrent neural network (RNN)-based deep learning architecture. It is primarily used for processing and predicting time series data or sequential data. Because this research focus on the long-term prediction, therefore we use LSTM. It overcomes the limitations of traditional RNNs, which struggle with learning long-term dependencies due to problems like the vanishing gradient.

% \begin{figure*}[t]
% \centering
% \includegraphics[height=2.4in, width=0.85\linewidth]{img/arch.png}
% \caption{The proposed model}
% \label{fig:model}
% \end{figure*}

% \subsection{Congestion evaluation}

%Quantity (the number of vehicles) and Velocity (average speed) between Tsurugashima JCT and Kawagoe IC on April 2022. The time interval of this dataset is 5 minutes. If the plotted point belongs to the orange area, it is considered to be congested.

\section{Experiment and Evaluation}\label{Experiment} 

\subsection{Experimental Setting and Evaluation Metrics}
All experiments are trained and tested on a linux server (CPU: 128 AMD EPYC 7702P 64-Core Processor/128 thread 2.0 GHz; GPU: NVIDIA RTX A6000 with 48GB memory*4). All deep neural networks were implemented using Pytorch. The random number generation seed was fixed to 1100 for experimental reproducibility. All the input data of $X$ was normalized using standard normalization, while the target $y$ value of speed was normalized by min-max normalization. 

We chose the following metrics to measure the performance of the models: mean absolute error (MAE, Equation~\ref{eqn:mae}). Here, the MAE was the result recovered from the normalized values, showing intuitive performance comparisons. The traffic speed value is the target of the prediction.

% , root mean square error (RMSE, Equation~\ref{eqn:rsme}), mean absolute percentage error (MAPE Equation~\ref{eqn:mape})
\begin{equation}
\centering
MAE = \frac{1}{n}\sum_{i=1}^{n}|\hat{y}-y_{i}|
\label{eqn:mae}
\end{equation}

% \begin{equation}
% \centering
% RSME = \sqrt{\frac{1}{n}\sum_{i=1}^{n} (\hat{y}-y_{i})^{2}}
% \label{eqn:rsme}
% \end{equation}

% \begin{equation}
% \centering
% MAPE = \frac{1}{n}\sum_{i=1}^{n}\frac{|\hat{y}-y_{i}|}{y_{i}}
% \label{eqn:mape}
% \end{equation}
% \subsection{Ablation Study}
\subsection{Ablation Study}
We progressively conducted two experiments step by step for effectively filtering parameter settings.
At first, an ablation study based on a relatively small amount of coarse-grained data of road E14 has been conducted. Table~\ref{tab:tateyama} summarizes the data details used in the ablation study. The input data is with 5-minute granularity and output data is with 1-hour granularity. To better examine the effectiveness of various features, we first conducted an ablation study. We compared the prediction results of various combinations of the features. 
\begin{table}[h]%11*7
\centering
\caption{Statistics of the Dataset}
\begin{tabular}{ll}
\toprule
Items& Values\\\hline
Road&  E14\\
Road segments&	16 IC segments\\
Total records & 157248\\
Total length&55.7 km\\
Time granularity of input& 5 minutes\\
Time granularity of output&  1 hour\\
Target Time & Apr. 3, 2021 - Sep. 30, 2022 (546 days)\\
Time length of Train Data & 434 days\\
Time length of Test Data & 112 days\\
\bottomrule
\end{tabular}
\label{tab:tateyama}
\end{table}

Table~\ref{tab:aba1} and Table~\ref{tab:aba2} show the results of the ablation study. Both evaluations used the input data of previous 24 hours. Table~\ref{tab:aba1} presents the prediction for the next day, while Table~\ref{tab:aba2} presents the one-day result of the next week.

\begin{table}[h]
\centering
\caption{Predict the next \textbf{one-day} result}
\begin{tabular}{lllll}
\toprule
\multicolumn{4}{c}{X} &  Y\_speed\\\hline
   X\_traffic &    X\_time    &  X\_search   & X\_search\_unspec   &MAE\\\hline
   \checkmark &  \checkmark & \checkmark   & \checkmark & \textbf{2.415} \\
   % \checkmark  &     & \checkmark     &     & 2.790 \\
   % \checkmark  &     &     &   \checkmark   &2.890  \\
    % \checkmark &     \checkmark   &        &     &  2.944\\
   \checkmark &     \checkmark   &   \checkmark  &    &  2.586 \\
    \checkmark &  \checkmark   &     & \checkmark    &2.979 \\
 \checkmark    &     &   \checkmark   &  \checkmark    &3.040  \\
\bottomrule
\end{tabular}
\label{tab:aba1}
\end{table}
Both experiments show that using the input of all features can achieve the best results. Whereas, while predicting the next one-day result, as shown in Table~\ref{tab:aba1}, the MAE value of using merely specified time (denoted as X\_search) dropped to 2.790, and while the MAE value of non-time unspecified (X\_search\_unspec) dropped to 2.586. These obvious performance drops clearly demonstrates corporating both two types of search records can significantly improve traffic prediction. On the other hand, the features of time (denoted as X\_time) 
are also important for the prediction model, because the model learning without time feature shows the worst MAE value of 3.04 (Table~\ref{tab:aba1}) and 3.408 (Table~\ref{tab:aba2}) respectively.
\begin{table}[h]
\centering
\caption{Predict \textbf{the same weekday of the next week}}
\begin{tabular}{lllll}
\toprule
\multicolumn{4}{c}{X} &  Y\_speed\\\hline
   X\_traffic &    X\_time    &  X\_search   & X\_search\_unspec   &MAE\\\hline
   \checkmark &  \checkmark   & \checkmark    & \checkmark    & \textbf{2.454} \\
   \checkmark &  \checkmark   &   \checkmark  &    & 2.492 \\
   \checkmark  & \checkmark    &      &  \checkmark   & 2.510 \\
   % \checkmark  &     &     &   \checkmark   &3.271  \\
   % \checkmark  & \checkmark    &     &      &3.291  \\
   \checkmark &      &   \checkmark     &  \checkmark   &  3.408\\
 % \checkmark    &     &   \checkmark   &    &3.522  \\
\bottomrule
\end{tabular}
\label{tab:aba2}
\end{table}

Based on the results of the pilot study with coarse-grained data, we further conducted an evaluation on a larger dataset with fine-grained. Table~\ref{tab:sta} summarizes the fine-grained data details used in the research. We used datasets from two roads for evaluation, where each road segment is more fine-grained, measuring approximately 2 kilometers (denoted as a KP segment).

\begin{table}[h]%11*7
\centering
\caption{Statistics of the traffic speed sub-dataset Items}
\begin{tabular}{ll}
\toprule
Items& Values\\\hline
Roads&  2; road code of E17 and E4\\
Road segments&	286 of KP segments\\
Total records & 12422592\\
Total length&285 km\\
Time granularity of input& 5 minutes\\
Time granularity of output&  1 hour\\
Target Time & Apr 3, 2021 - Jan 31, 2024 (1006 days)\\
Time length of Train Data & 782 days\\
Time length of Validation Data & 112 days\\
Time length of Test Data & 112 days\\
\bottomrule
\end{tabular}
\label{tab:sta}
\end{table}

\subsection{Evaluation Result}
Based on the result of the ablation study, we utilized all the features for evaluation. The input data is with 5-minute granularity and output data is with 1-hour granularity.  As listed in Table~\ref{tab:set}, we evaluated the following evaluation tasks, comparing the results for different values of input data size, prediction intervals (e.g., predicting the outcome for the next day or the same weekday of the following week), and the output data size. The goal was to determine the best combination of these parameters.

\begin{table}[h]
\centering
\caption{Parameters setting for evaluation comparison}
\begin{tabular}{l|l|l}
\toprule
            Parameters            & Values &  Description\\\hline
\multirow{2}{*}{Input\_size}      & 24*12 & Input one day's data \\
                                  & 168*12 & Input one week's data \\
                                  \cline{1-3} 
\multirow{2}{*}{N\_day\_interval} &  0&  Prediction of next day\\
                                  &  6&  Prediction of the same weekday of next week \\\cline{1-3} 
Output\_size    &  24&  Output of one day's result\\
                                  % &  168 &  Output of one week's result\\
\bottomrule
\end{tabular}
\label{tab:set}
\end{table}

Table~\ref{tab:results} summarizes the performance comparison results for the  model tested on different roads: E4 and E17. For the test datasets, the input sizes are either 24 (one-day) or 168 (one-week), and the output size is fixed at 24 (one day). The table also shows day intervals of either 0 (denotes the prediction of the next day) or 6 (denotes the prediction of the same weekday of the next week) with the Mean Absolute Error (MAE) for each scenario.
\begin{itemize}
\item  For the road of E4: The MAE ranges from 3.354 to 3.904, with smaller input sizes (one-day data) generally leading to lower MAE values.
\item For the road of E17: The MAE ranges from 3.596 to 3.902, with similar patterns where the smaller input size of one-day results in slightly better performance than the larger input size of 168.
\end{itemize}

In both target roads, \textbf{input patterns of one-day data to predict the result of the next day achieves the best result}. 0-day interval results in slightly better performance than the 6-day interval for the same input size.

% \begin{itemize}
% \item Comparison of different size of past data used for learning. To examine the difference of all (more than 3 years) or partial (e.g. the last 2 years or 1 year).
% \item Comparison of results of different values of input data size, interval (i.g., to predict the result of next day or the day of next week), and output data size. To examaine the best combination pattern.
% \end{itemize}

%Our result can not only contribute to more precise railway passenger count prediction,but also provide an insight to model design for task on urban railway network.

\begin{table}[h]
\centering
\caption{Results of different input size, interval\_days and output size}
\begin{tabular}{lllll}
\toprule
Road & In\_size & n\_day\_interval & out\_size & MAE of speed\\\hline
\multirow{4}{*}{ E4} &      24    &         0         &       24    &    \textbf{3.354}\\
  &    24  &          6      &        24 & 3.367   \\
   &    168    &          0     &        24 & 3.868   \\
&    168    &          6      &        24 & 3.904   \\\hline
\multirow{4}{*}{ E17}  &    24      &          0    &        24 & \textbf{3.596} \\
 &    24      &          6   &        24 & 3.618   \\
 &    168     &          0    &        24 & 3.868  \\
  &    168     &          6   &        24 & 3.902  \\
\bottomrule
\end{tabular}
\label{tab:results}
\end{table}

% \subsection{Discussion}

\section{Conclusion and Future Work}
%Our result can not only contribute to more precise railway passenger count prediction, but also provide an insight to model design for task on urban railway network.

This paper presented a novel model design for federating route search records for improving long-term traffic prediction. The proposed model is  compatible with different data features and time granularities and lengths. It was practically evaluated using the latest real-world data in the collaboration with NEXCO East Japan. We also introduced a process of route search records and the analysis results show that these search records effectively capture trends in traffic status. Notably, time specified records exhibit a strong correlation with future traffic volume.
The model was trained on more than two years of data, and the average speeds were predicted target. The ablation study demonstrated that incorporating all features, including the search count as input, achieved the best performance.

Additionally, we examined the results from multiple perspectives related to parameter settings within the time dimension. We evaluated varying input lengths, such as historical data from one day or one weekday, as well as different time intervals, including predictions for the next day or the same weekday of the following week. The results indicate that the best performance was achieved in most cases by using historical data to predict the outcome for the following day. Future research could explore applying this federation architecture to other feature dimensions, such as weather data, which could further enhance the effectiveness of our proposed model.

% Furthermore, we examined the results from multiple perspectives on the time dimension. We evaluated varying input lengths, such as one day or one weekday of historical data, as well as different time intervals. The results show that performance was maximized in most cases by using search data from one day before the prediction time and predicting the outcome for the following day.

% In fact, the importance of the search log data was
% found to be equivalent to that of temporal features, and its
% contribution to traffic congestion prediction was significant.
% We believe that utilizing the search log data will allow us to
% determine the future road conditions several days in advance,
% leading to less traffic congestion and the development of ITS

% Besides, we also investigated the interprebability of the models with the corresponding weight graphs. As the pioneering effort to employ the graph networks learning on diverse and heterogeneous weight graphs for passenger count on the large scale railway network, we believe our work could inspire and provide references for the followers to develop similar or more novel research.
% Together with utilizing the online search logs, the prediction has been improved.

\section{Acknowledgement}
This study was conducted under a joint research project between the University of Tokyo and the East Nippon Expressway Co., Ltd. (NEXCO East). The data used in this study, such as historical traffic data, online search logs, and road structure information, were provided by NEXCO East. We gratefully acknowledge the kind support provided by the NEXCO East.

\balance
\bibliography{IEEEabrv,aaai22}

\begin{thebibliography}{10}
\providecommand{\url}[1]{#1}
\csname url@rmstyle\endcsname
\providecommand{\newblock}{\relax}
\providecommand{\bibinfo}[2]{#2}
\providecommand\BIBentrySTDinterwordspacing{\spaceskip=0pt\relax}
\providecommand\BIBentryALTinterwordstretchfactor{4}
\providecommand\BIBentryALTinterwordspacing{\spaceskip=\fontdimen2\font plus
\BIBentryALTinterwordstretchfactor\fontdimen3\font minus
  \fontdimen4\font\relax}
\providecommand\BIBforeignlanguage[2]{{%
\expandafter\ifx\csname l@#1\endcsname\relax
\typeout{** WARNING: IEEEtran.bst: No hyphenation pattern has been}%
\typeout{** loaded for the language `#1'. Using the pattern for}%
\typeout{** the default language instead.}%
\else
\language=\csname l@#1\endcsname
\fi
#2}}

\bibitem{tian2015predicting}
Y.~Tian and L.~Pan, ``Predicting short-term traffic flow by long short-term
  memory recurrent neural network,'' in \emph{2015 IEEE international
  conference on smart city/SocialCom/SustainCom (SmartCity)}.\hskip 1em plus
  0.5em minus 0.4em\relax IEEE, 2015, pp. 153--158.

\bibitem{ma2015long}
X.~Ma, Z.~Tao, Y.~Wang, H.~Yu, and Y.~Wang, ``Long short-term memory neural
  network for traffic speed prediction using remote microwave sensor data,''
  \emph{Transportation Research Part C: Emerging Technologies}, vol.~54, pp.
  187--197, 2015.

\bibitem{yang2019mf}
D.~Yang, S.~Li, Z.~Peng, P.~Wang, J.~Wang, and H.~Yang, ``Mf-cnn: Traffic flow
  prediction using convolutional neural network and multi-features fusion,''
  \emph{IEICE TRANSACTIONS on Information and Systems}, vol. 102, no.~8, pp.
  1526--1536, 2019.

\bibitem{ma2017learning}
X.~Ma, Z.~Dai, Z.~He, J.~Ma, Y.~Wang, and Y.~Wang, ``Learning traffic as
  images: A deep convolutional neural network for large-scale transportation
  network speed prediction,'' \emph{Sensors}, vol.~17, no.~4, p. 818, 2017.

\bibitem{ye2020build}
J.~Ye, J.~Zhao, K.~Ye, and C.~Xu, ``How to build a graph-based deep learning
  architecture in traffic domain: A survey,'' \emph{IEEE Transactions on
  Intelligent Transportation Systems}, 2020.

\bibitem{yin2021deep}
X.~Yin, G.~Wu, J.~Wei, Y.~Shen, H.~Qi, and B.~Yin, ``Deep learning on traffic
  prediction: Methods, analysis, and future directions,'' \emph{IEEE
  Transactions on Intelligent Transportation Systems}, vol.~23, no.~6, pp.
  4927--4943, 2021.

\bibitem{zhang2018predicting}
J.~Zhang, Y.~Zheng, D.~Qi, R.~Li, X.~Yi, and T.~Li, ``Predicting citywide crowd
  flows using deep spatio-temporal residual networks,'' \emph{Artificial
  Intelligence}, vol. 259, pp. 147--166, 2018.

\bibitem{ryu2019intelligent}
S.~Ryu and D.~Kim, ``Intelligent highway traffic forecast based on deep
  learning and restructured road models,'' in \emph{2019 IEEE 43rd Annual
  Computer Software and Applications Conference (COMPSAC)}, vol.~2.\hskip 1em
  plus 0.5em minus 0.4em\relax IEEE, 2019, pp. 110--114.

\bibitem{zhang2017deep}
J.~Zhang, Y.~Zheng, and D.~Qi, ``Deep spatio-temporal residual networks for
  citywide crowd flows prediction,'' in \emph{Proceedings of the AAAI
  conference on artificial intelligence}, vol.~31, no.~1, 2017.

\bibitem{nexco}
``Nexco east,'' \url{https://www.e-nexco.co.jp/en/}, accessed Oct-2024.

\bibitem{okutani1984dynamic}
I.~Okutani and Y.~J. Stephanedes, ``Dynamic prediction of traffic volume
  through kalman filtering theory,'' \emph{Transportation Research Part B:
  Methodological}, vol.~18, no.~1, pp. 1--11, 1984.

\bibitem{williams2003modeling}
B.~M. Williams and L.~A. Hoel, ``Modeling and forecasting vehicular traffic
  flow as a seasonal arima process: Theoretical basis and empirical results,''
  \emph{Journal of transportation engineering}, vol. 129, no.~6, pp. 664--672,
  2003.

\bibitem{chandra2009predictions}
S.~R. Chandra and H.~Al-Deek, ``Predictions of freeway traffic speeds and
  volumes using vector autoregressive models,'' \emph{Journal of Intelligent
  Transportation Systems}, vol.~13, no.~2, pp. 53--72, 2009.

\bibitem{leshem2007traffic}
G.~Leshem and Y.~Ritov, ``Traffic flow prediction using adaboost algorithm with
  random forests as a weak learner,'' \emph{International Journal of
  Mathematical and Computational Sciences}, vol.~1, no.~1, pp. 1--6, 2007.

\bibitem{may2008vector}
M.~May, D.~Hecker, C.~K{\"o}rner, S.~Scheider, and D.~Schulz, ``A
  vector-geometry based spatial knn-algorithm for traffic frequency
  predictions,'' in \emph{2008 IEEE International Conference on Data Mining
  Workshops}.\hskip 1em plus 0.5em minus 0.4em\relax IEEE, 2008, pp. 442--447.

\bibitem{fu2016vehicle}
H.~Fu, H.~Ma, Y.~Liu, and D.~Lu, ``A vehicle classification system based on
  hierarchical multi-svms in crowded traffic scenes,'' \emph{Neurocomputing},
  vol. 211, pp. 182--190, 2016.

\bibitem{zhao2017lstm}
Z.~Zhao, W.~Chen, X.~Wu, P.~C. Chen, and J.~Liu, ``Lstm network: a deep
  learning approach for short-term traffic forecast,'' \emph{IET Intelligent
  Transport Systems}, vol.~11, no.~2, pp. 68--75, 2017.

\bibitem{wu2016short}
Y.~Wu and H.~Tan, ``Short-term traffic flow forecasting with spatial-temporal
  correlation in a hybrid deep learning framework,'' \emph{arXiv preprint
  arXiv:1612.01022}, 2016.

\bibitem{li2017diffusion}
Y.~Li, R.~Yu, C.~Shahabi, and Y.~Liu, ``Diffusion convolutional recurrent
  neural network: Data-driven traffic forecasting,'' \emph{arXiv preprint
  arXiv:1707.01926}, 2017.

\bibitem{cheng2018deeptransport}
X.~Cheng, R.~Zhang, J.~Zhou, and W.~Xu, ``Deeptransport: Learning
  spatial-temporal dependency for traffic condition forecasting,'' in
  \emph{2018 International Joint Conference on Neural Networks (IJCNN)}.\hskip
  1em plus 0.5em minus 0.4em\relax IEEE, 2018, pp. 1--8.

\bibitem{ge2024k}
H.~Ge, T.~Michikata, and N.~Koshizuka, ``K-neighboring on multi-weighted graphs
  for passenger count prediction on railway networks,'' \emph{Journal of
  Information Processing}, vol.~32, pp. 575--585, 2024.

\bibitem{hangli2022multi}
G.~Hangli, L.~Lin, R.~Jiang, T.~Michikata, and N.~Koshizuka, ``Multi-weighted
  graphs learning for passenger count prediction on railway network,'' in
  \emph{2022 IEEE 46th Annual Computers, Software, and Applications Conference
  (COMPSAC)}.\hskip 1em plus 0.5em minus 0.4em\relax IEEE, 2022, pp. 374--382.

\bibitem{pems}
PeMS, ``Performance measurement system (pems),''
  \url{https://dot.ca.gov/programs/traffic-operations/mpr/pems-source},
  accessed Oct-2024.

\bibitem{q-traffic}
Q-traffic, ``Q-traffic,'' \url{https://github.com/JingqingZ/BaiduTraffic},
  accessed Oct-2024.

\bibitem{seattle-loop}
``Loop dataset,'' \url{https://github.com/zhiyongc/Seattle-Loop-Data}, 2023,
  accessed Oct-2024.

\bibitem{kosugi2022traffic}
Y.~Kosugi, I.~Matsunaga, H.~Ge, T.~Michikata, and N.~Koshizuka, ``Traffic
  congestion prediction using toll and route search log data,'' in \emph{2022
  IEEE International Conference on Big Data (Big Data)}.\hskip 1em plus 0.5em
  minus 0.4em\relax IEEE, 2022, pp. 5971--5978.

\bibitem{matsunaga2023improving}
I.~Matsunaga, Y.~Kosugi, G.~Hangli, T.~Michikata, and N.~Koshizuka, ``Improving
  long-term traffic prediction with online search log data,'' in \emph{2023
  IEEE 47th Annual Computers, Software, and Applications Conference
  (COMPSAC)}.\hskip 1em plus 0.5em minus 0.4em\relax IEEE, 2023, pp.
  1750--1755.

\bibitem{dorapura}
NEXCO-East-Japan, ``Dorapura service,'' \url{https://en.driveplaza.com/}, 2023,
  accessed Oct-2024.

\end{thebibliography}

\bibliographystyle{IEEEtran}
%\bibliography{IEEEabrv,Bibliography}

\end{document}